# Interpretable estimation of the risk of heart failure hospitalization from a 30-second electrocardiogram

Sergio González[*], Wan-Ting Hsieh[*], Davide Burba[*], Trista Pei-Chun Chen[*],
Chun-Li Wang[†‡], Victor Chien-Chia Wu[†‡], Shang-Hung Chang[†§¶]

[*]*AI Center, Inventec Corporation,* Taipei, Taiwan
gonzalez-vazquez.sergio@inventec.com
[†]*Cardiovascular Division, Department of Internal Medicine, Chang Gung Memorial Hospital,* Linkou, Taiwan
[‡]*College of Medicine, Chang Gung University,* Taoyuan, Taiwan
[§]*Center for Big Data Analysis and Statistics, Chang Gung Memorial Hospital,* Linkou, Taiwan
[¶]*Graduate Institute of Nursing, Chang Gung University of Science and Technology,* Taoyuan, Taiwan

*Abstract*—Survival modeling in healthcare relies on explainable statistical models; yet, their underlying assumptions are often simplistic and, thus, unrealistic. Machine learning models can estimate more complex relationships and lead to more accurate predictions, but are non-interpretable. This study shows it is possible to estimate hospitalization for congestive heart failure by a 30 seconds single-lead electrocardiogram signal. Using a machine learning approach not only results in greater predictive power but also provides clinically meaningful interpretations. We train an eXtreme Gradient Boosting accelerated failure time model and exploit SHapley Additive exPlanations values to explain the effect of each feature on predictions. Our model achieved a concordance index of 0.828 and an area under the curve of 0.853 at one year and 0.858 at two years on a held-out test set of 6,573 patients. These results show that a rapid test based on an electrocardiogram could be crucial in targeting and treating high-risk individuals.

## I. Introduction

The application of machine learning (ML) to the healthcare industry has received immense interest in the past few years due to the medical and economical positive impact. However, ML health models need to be interpretable and understandable since a wrong estimation could negatively affect human lives. Interpretability is a hot topic in the ML community [1], with many novel methods and algorithms having been proposed under two main alternatives. First, the default choice in classical medical applications is an inherent explainable model, such as linear models. However, they make strong and often unrealistic assumptions about the functional form of processes. The second alternative disentangles the interpretation process from the model. In this setting, the model can take an arbitrary form, thus allowing us to estimate the underlying process more realistically and accurately, while still being able to explain the predictions via interpretation methods. In particular, SHAP (SHapley Additive exPlanations) [2] allows decomposing a prediction of any model into the sum of the contribution of all features and the prediction expectation. It is the only method able to provide a coherent explanation both at a local level and at a global level.

In this paper, we focus on interpretable survival analysis using ML models. Specifically, we tackle estimating hospitalization for congestive heart failure (CHF) based on only 30 seconds of a single-lead electrocardiogram (ECG) signal and basic demographic and clinical information. CHF is a common and highly lethal syndrome whose symptoms and signs are caused by cardiac dysfunction [3]. The standard procedure to estimate CHF development relies on the Cox model [4] and typically requires 24 hours of ECG data [5] or depends on other information, such as blood test analyses or drug intakes [6]. However, due to the linear assumption of a covariate and hazard function in the Cox model, it cannot capture covariate interaction. Instead, we use an eXtreme Gradient Boosting (XGBoost) accelerated failure time (AFT) model [7] that allows non-linear regression. Moreover, we include a SHAP kernel to explain model behavior and the decision of a single prediction. Our contribution to the literature is two-fold. First, we demonstrate that it is possible to accurately estimate the development of CHF by using only 30 seconds of an ECG. Second, we show that an ML model paired with SHAP values not only improves the prediction accuracy, but also provides an interpretation that is clinically realistic.

## II. Survival models

Survival models are a particular class of regression models that deal with time-to-event data [4, 8]. They learn the time elapsed before the occurrence of an event from uncensored samples, and leverage the censored data for which the event never happened during the observation period. Among survival models, the Cox model is the most widely used, assuming a functional form for the hazard function [4] (i.e., the instantaneous failure rate) as $\lambda(t|x) = \lambda_0(t)e^{\beta^T x}$, where x denotes a vector of features, β is a vector of feature coefficients, and $\lambda_0(t)$ is a positive function called "baseline hazard". Since the Cox model is a multiple linear regression, it is easy to interpret the relationship between the features and hazard ratio; yet, this has some limitations. First, the model does not take into account feature interaction. Second, the features' effect is constant over time, since the time dependency is completely encoded in the baseline hazard. Several ML-based approaches have been proposed to relax the restrictive hypothesis of the Cox model, such as XGBoost [7] and DeepHit [9].

## III. INTERPRETABLE CHF ESTIMATION FRAMEWORK

Our interpretable CHF survival framework first processes the ECG signal to extract a reduced and interpretable set of features. These features are combined with demographic and clinical data from the subjects. Then, we incorporate the XGBoost AFT model to predict the probability of the future development of CHF. Finally, our framework leverages SHAP values to interpret the global model behavior and each individual estimation.

### A. Data collection

The ECG data were collected at Chang Gung Memorial Hospital (Linkou, Taiwan) using a three-lead holter machine for 24 hours between 01/05/2009 and 12/29/2017. The patient cohort comprised 21,891 subjects, of which 50.4% were men and 49.6% women. Patients were monitored for CHF until 03/06/2020, for a minimum period of two years and two months. A total of 1,805 (8.2%) patients were hospitalized for CHF over the observation period, with 919 (4.2%) within one year and 1,195 (5.5%) within two years. The diagnosis of CHF was based on their hospitalization with a primary discharge diagnosis of CHF. The censoring time was set to 03/06/2020 if the patient did not have CHF until that date; we did not consider death or other clinical events as censoring.

### B. ECG processing and features extraction

We used the second channel ECG signal from each patient's first visit. Each recording was split into segments of 30 seconds, denoised by applying a band-pass Butterworth filter between 0.5 and 45 Hz, and scaled between 0 and 1. After that, we extracted the location of the R-peaks using the Hamilton algorithm [10]. Centered at each R-peak location, we identified each heart cycle with a duration equal to the minimum length of the two RR intervals before and after the current R-peak. We then resampled each cycle to a length of 100 via linear interpolation, and computed the segment quality as the average correlation between each cycle and the average across cycles. Finally, after discarding segments with quality below 0.85 (8% of the total), we randomly selected one 30-second segment for each patient.

Next, we extracted two types of interpretable EGC features. (1) HRV features based on the timing of the R-peaks. (2) Shape features characterize the ECG average heart cycle. Specifically, we computed the timing and amplitudes of the P, Q, S, and T waves in the extracted mean cycle; we did not consider the R wave because cycles are centered on it and normalized. ECG wave identification is considerably hard since the waves could be subtle, missing, or even inverted. Therefore, for each wave, the most prominent peak or valley is selected from the available candidates in its specific cycle region. Fig. 1 show a sample of two average heart cycles, along with the extracted wave components. Bottom-right patient exhibits an inverted T wave.

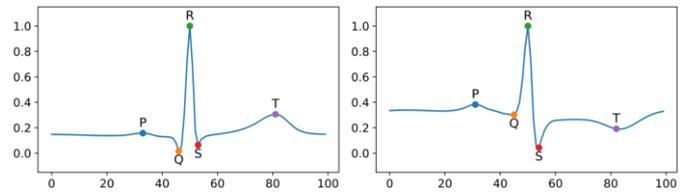

Figure 1. Average heart cycles and the extracted wave features.

These ECG features were paired with patients' demographic and clinical basic information. Each numeric feature was normalized using a quantile transformation. Such normalization alleviated the potential problems due to outliers and skewed distributions and simplified interpretability. A complete overview and description of all features are given in Table I. Waves timing are in the range from 0 to 100 while the amplitudes are from 0 to 1, due to the resampling and normalization of each mean cycle. P_elev and s_elev are discarded from our dataset due to high Pearson's correlations with t_elev and q_elev, respectively.

TABLE I.  SUBJECT FEATURES OVERVIEW

|  | Abbr. | % or mean±std |
|---|---|---|
| **Demographics** | | |
| Age at date of ECG | age | 58.0±19.6 |
| Sex (male 1 / female 0) | sex | 50.4% |
| **ECG - HRV** | | |
| Mean heart rate | mean_hr | 75.0±18.0 |
| Std. Of RR intervals | sdnn | 0.1±0.1 |
| Ratio of std. of Poincare analysis [11] | ratio_sd1_sd2 | 0.8±1.7 |
| **ECG - Shape** | | |
| P wave timing | p_timing | 24.9±11.5 |
| Q wave timing | q_timing | 44.1±3.4 |
| S wave timing | s_timing | 55.7±3.7 |
| T wave timing | t_timing | 85.5±7.0 |
| Q wave amplitude | q_amplitude | 0.1±0.2 |
| T wave amplitude | t_amplitude | 0.4±0.2 |
| **Clinical history (diagnosed diseases)** | | |
| Atrial fibrillation | AF_history | 9.5% |
| Chronic kidney disease | CKD_history | 11.5% |
| Chronic obstructive pulmonary disease | COPD_history | 11.7% |
| Diabetes mellitus | DM_history | 17.9% |
| Hyperlipidemia | HL_history | 22.4% |
| Hypertension | HTN_history | 37.9% |
| Ischemic heart disease | IHD_history | 17.7 % |
| Myocardial infarction | MI_history | 3.2% |
| Stroke | STROKE_history | 9.4% |
| Valvular heart disease | VHD_history | 6.9% |

### C. Modeling and interpretation

XGBoost AFT [7] is a powerful gradient boosting algorithm for supervised that has been adapted to survival setting by generalizing the so-called "accelerated failure time" model [4]. Specifically, the model assumes that $\ln(T) = \tau(x) + \epsilon$, where T is the time-to-event random variable, $\tau(x)$ is estimated by the gradient boosting engine, and $\epsilon$ is an error term. In our application, we assume a log-logistic (Fisk) distribution of the time-to-event. $\epsilon$ follows a zero-mean logistic distribution parametrized by a standard deviation, the hyperparameter σ. Therefore, we can easily convert the XGBoost's output to a predicted distribution. The model is trained using a negative log-likelihood loss function. As our dataset is imbalance, with only 8.2% having CHF, we apply instance weighting during training as well as introduce the L1 and L2 regularization.

We leveraged SHAP tool [2] to enable the global and local interpretation of the model and its predictions. The global interpretation is given by feature importance and the general effect on model output computed by the mean of contributions. Moreover, the local interpretation of a given prediction is supported, by showing how each concrete feature value contributes to model output. Two implementations have been used, namely TreeSHAP and KernelSHAP. Due to its high efficiency, our framework used TreeSHAP to obtain a global interpretation of the XGBoost model behavior in the model output space (i.e., the expected logarithm of the timeto-event). For the individual interpretation, we applied the KernelSHAP for decomposition in the probability space.

## IV. RESULTS

### A. CHF prediction performance

The 21,891 30-second ECG segments from unique patients were split into 70% training and 30% test sets. We compared the XGBoost AFT model with the Cox model and with a stateof-the-art survival model, DeepHit [9]. The prediction performances were evaluated using Antolini's concordance index (C-index) [12] and the areas under the ROC curves (AUC) within one and two years. The discriminatory power at all time horizons was tested by the average cumulative/dynamic AUC (c/d AUC) [13]. We trained the models via 5-fold cross validation with the training set, and selected the hyperparameters with the highest validation C-index. Then, we computed the final performance on the test set. Table II shows the performance results and the 90% confidence intervals of the models. Overall, XGBoost outperforms the other two models with the highest scores, and narrowest confidence intervals. XGBoost maintained the highest discriminatory power at the different horizons with a c/d AUC of 0.849. The Cox model has the lowest performance, and shows a notable margin compared to XGBoost

TABLE II. PERFORMANCE FOR CHF PREDICTION ON THE TEST SET

|  | XGBoost | DeepHit | Cox |
| --- | --- | --- | --- |
| C-index | **0.828** [0.816, 0.840] | 0.809 [0.795, 0.822] | 0.807 [0.794, 0.820] |
| AUC 1 year | **0.853** [0.837, 0.868] | 0.826 [0.807, 0.845] | 0.824 [0.805, 0.842] |
| AUC 2 years | **0.858** [0.844, 0.871] | 0.836 [0.819, 0.852] | 0.834 [0.817, 0.850] |
| Avg. c/d AUC | **0.849** [0.835, 0.862] | 0.826 [0.810, 0.841] | 0.822 [0.806, 0.837] |

### B. Increasing ECG segment length

We aimed to evaluate the limitations of using 30-second ECGs over longer segments. Therefore, we repeated the experiments with different lengths (i.e., from 1 to 15 minutes and 24 hours). Fig. 2 shows their corresponding C-indices and the confidence intervals computed via bootstrapping. Although using longer segments generally leads to an improvement of the C-index, this improvement is limited. For the XGBoost model, using 30 times more data (i.e., 15 minutes) or 24- hour recordings improve the C-index by roughly 1%, while considerably increasing the costs of data collection and inspection. This improvement would be more significant (around 2%) under the Cox model, meaning that the XGBoost model is more robust to segment length. Moreover, the basic XGBoost with 30-second segments still outperforms the Cox model with 15-minute data and is comparable to the 24-hour model.

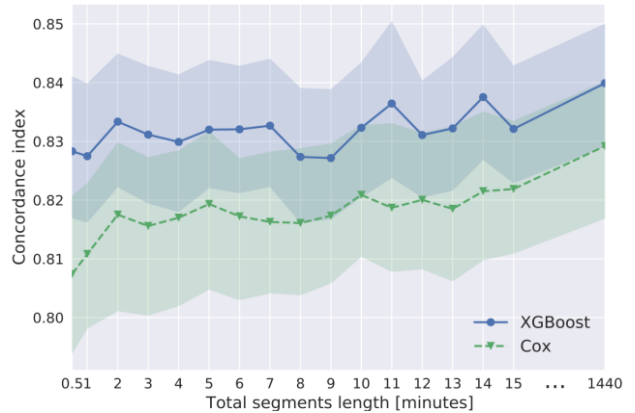

Figure 2. C-index for the test set for different total segments length.

### C. Interpretation

Fig. 3 shows the global impact of the 10 most important features on model predictions, where the feature and SHAP values are on the horizontal and vertical axes, respectively. Here, negative SHAP values correspond to shorter time-to-events (i.e., higher risks) and vice versa. Note that multiple features show a non-monotonic trend, which would be impossible to estimate using a Cox model.

Among all features, *t_amplitude* is the most important one. Lower values of *t_amplitude* lead to shorter time-to-events (i.e., higher predicted risks). These correspond to inverted T waves, which are associated with several CHF risk factors and abnormalities [14], such as myocardial ischemia [15], bundle branch block [16], and premature ventricular contraction [17]. Hyper-acute T waves, which are a potential warning of myocardial infarction [18], exhibit a decreasing contribution. The second most important feature is the HRV *ratio_sd1_sd2*. Overall, its SHAP plot shows a non-monotonic descending trend. That is, high values lead to shorter predicted times. This agrees with previous findings [19]. Lastly, the age variable shows a decreasing trend, as expected.

Most of the disease variables lead to higher predicted risks, as expected. The most important ones are CKD history, DM history and IHD history. These are consistent with recent medical findings, as there is high prevalence of heart failure in patients with CKD [20] or DM [21]. Besides, ischemia is also a major underlying pathogenic factor in CHF [15]. Finally, we observed that men tend to have a higher risk than women.

Aside from the general model behavior, Fig. 4 shows an example of a single patient at the inference time. The model predicts that a patient will develop CHF within one year with a probability of 19.94%. The major reason comes from an inverted T wave (*t_amplitude* = 5.3% percentile), followed by diagnosed myocardial infarction and diabetes mellitus. A normal T timing (*t_timing* = 22% percentile) and a normal HRV (*ratio_sd1_sd2* = 30.6% percentile) decrease the risk.

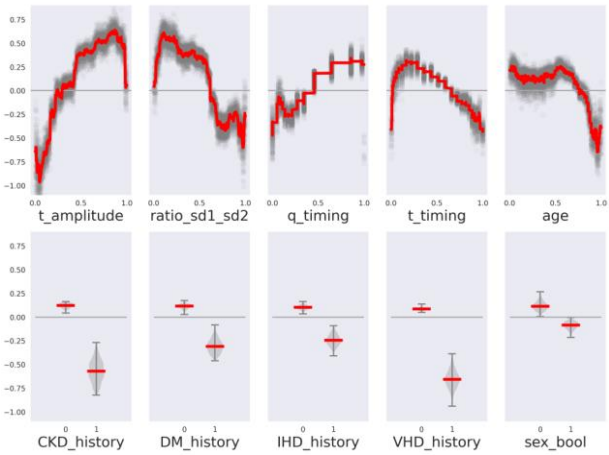

Figure 3. SHAP values for the entire test set. Numeric features are shown at the top and binary features at the bottom. The red thick lines represent the SHAP moving medians.

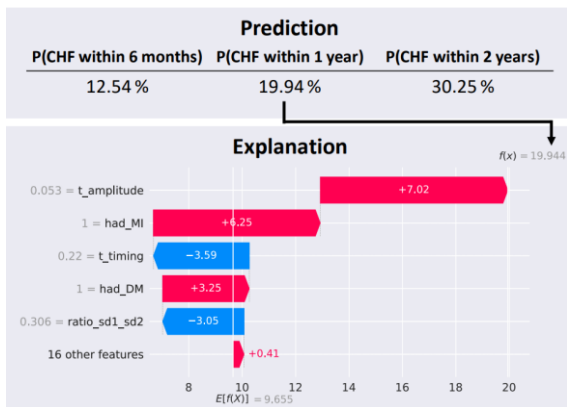

Figure 4. Model prediction and explanation for a patient who developed CHF 3.9 years after study enrollment.

## V. CONCLUSION

This study shows that it is possible to estimate CHF hospitalization using only a 30-second ECG. Moreover, our ML framework based on XGBoost and SHAP allowed us to achieve better performances and more realistic interpretations. In addition, a 30-second ECG was enough for our model to reach a high predictive performance comparable to the 24-hour ECG model. The explanations for the predictions of our model are consistent with the medical literature. One of the future direction would be to explore the interpretability of the automatic feature extraction models, for capturing additional risk factors from the ECG signal.